# A Ray-based Approach for Boundary Estimation of Fiber Bundles Derived from Diffusion Tensor Imaging

M. H. A. Bauer[1,3], S. Barbieri[2], J. Klein[2], J. Egger[1,3], D. Kuhnt[1], B. Freisleben[3], H.-K. Hahn[2], Ch. Nimsky[1]

[1] Department of Neurosurgery, University of Marburg, Marburg, Germany
[2] Fraunhofer MEVIS - Institute for Medical Image Computing, Bremen, Germany
[3] Department of Mathematics and Computer Science, University of Marburg, Germany

**Keywords**– *Diffusion Tensor Imaging · Fiber Tracking · Boundary Estimation · Ray-based Approach*

**Purpose**

Diffusion Tensor Imaging (DTI) is a non-invasive imaging technique that allows estimation of the location of white matter tracts in-vivo, based on the measurement of water diffusion properties. For each voxel, a second-order tensor can be calculated by using diffusion-weighted sequences (DWI) that are sensitive to the random motion of water molecules. Given at least 6 diffusion-weighted images with different gradients and one unweighted image, the coefficients of the symmetric diffusion tensor matrix can be calculated. Deriving the eigensystem of the tensor, the eigenvectors and eigenvalues can be calculated to describe the three main directions of diffusion and its magnitude [1-3]. Using DTI data, fiber bundles can be determined, to gain information about eloquent brain structures. Especially in neurosurgery, information about location and dimension of eloquent structures like the corticospinal tract or the visual pathways is of major interest. Therefore, the fiber bundle boundary has to be determined. In this paper, a novel ray-based approach for boundary estimation of tubular structures is presented.

**Methods**

DTI in combination with fiber tracking algorithms (e.g. deflection based) allows estimation of the position and course of fiber tracts in the human brain. To use this information for navigation in neurosurgical interventions, the 3D surface and 2D contours of arbitrarily views are necessary [4].

Depending on a manual, given seed region (2D contour), fiber tracking is applied to the data set. With the help of two regions of interest (ROI), the tracked bundle is restricted to the structure of interest. Between these two ROIs, the centerline of the bundle is derived as the starting point [5].

The given centerline is sampled at *n* points, and for each of the points a plane upright to the centerline direction given by the difference vector of two consecutive centerline points is computed. Within each plane, *k* equally distributed vectors around the centerline are calculated for the ray directions. Along these directions, *m* equally spaced points are computed with distance *d [mm]* between each of them.

Then, the tensor information is extracted for each evaluation point including the eigenvectors and eigenvalues as well as the fractional anisotropy (FA) value that describes the fraction of magnitude of the tensor that is ascribed to the anisotropic diffusion [6]. Depending on several threshold criteria concerning the FA value and two angle parameters, for each ray a boundary point is estimated. To prevent the surface from extreme outliers due to noise and artifacts, the 2D contours are explored stepwise by comparing the index of contour points of the same ray in neighbored layers. If the distance between them exceeds a user defined threshold, the actual point is corrected inwards (if it is too far away from the center) or outwards, respectively, until the distance criterion is fulfilled. For 3D surface construction, the point cloud is triangulated. The whole process of fiber bundle estimation is shown in Figure 1.

**Results**

To evaluate the method, software phantoms were used with known location and dimension of fiber tracts. Real patient data was not used for quality evaluation of the segmentation in a first step due to the missing ground truth to compare against. The Dice Similarity Coefficient (DSC) is used for quantifying the degree of overlap between the two segmented objects (ground truth and results from the presented method) [7].

The software phantom describes a portion of a torus with a diameter of *10mm* of the area cross-section with a voxel size of *1x1x1mm³*. Six gradient directions were used and complex Gaussian noise was applied.
Several parameters (number of samples along the centerline, number of rays per layer, distance between two points along a ray) were changed systematically.

Looking at the sampling points along the centerline, the DSC does not vary significantly between 33 and 65 sampling points. The number of directions for the given bundle has a greater impact on the result. Using at least 8 directions, the DSC changes only marginally. With fewer directions, the DSC is significantly worse. Looking at the distance between the evaluation points, the results do not vary much, but in combination with a low number of rays, the DSC decreases with smaller distances between the evaluation points (see Figure 2).

**Conclusion**

In this paper, a new ray-based approach for boundary estimation of tracked fiber bundles was presented. Starting with the centerline of a tracked fiber bundle, rays with sampled points are sent out radially in different cross-sectional planes along the centerline. With the help of some stopping criteria concerning the FA value, the angle between the main directions of neighbored sampled points and between the main directions of sampled points and the center point, the boundary of the fiber bundle is determined. With the use of several constraints – avoiding extreme outliers – a 3D surface of the fiber bundle is calculated and triangulated. Evaluated with several software phantoms – in order to have a ground truth to compare against – the method yielded a Dice Similarity Coefficient (DSC) between 74.7% and 91.5%. There are several areas of future work; for example, the method can be enhanced for bifurcated fiber bundles.

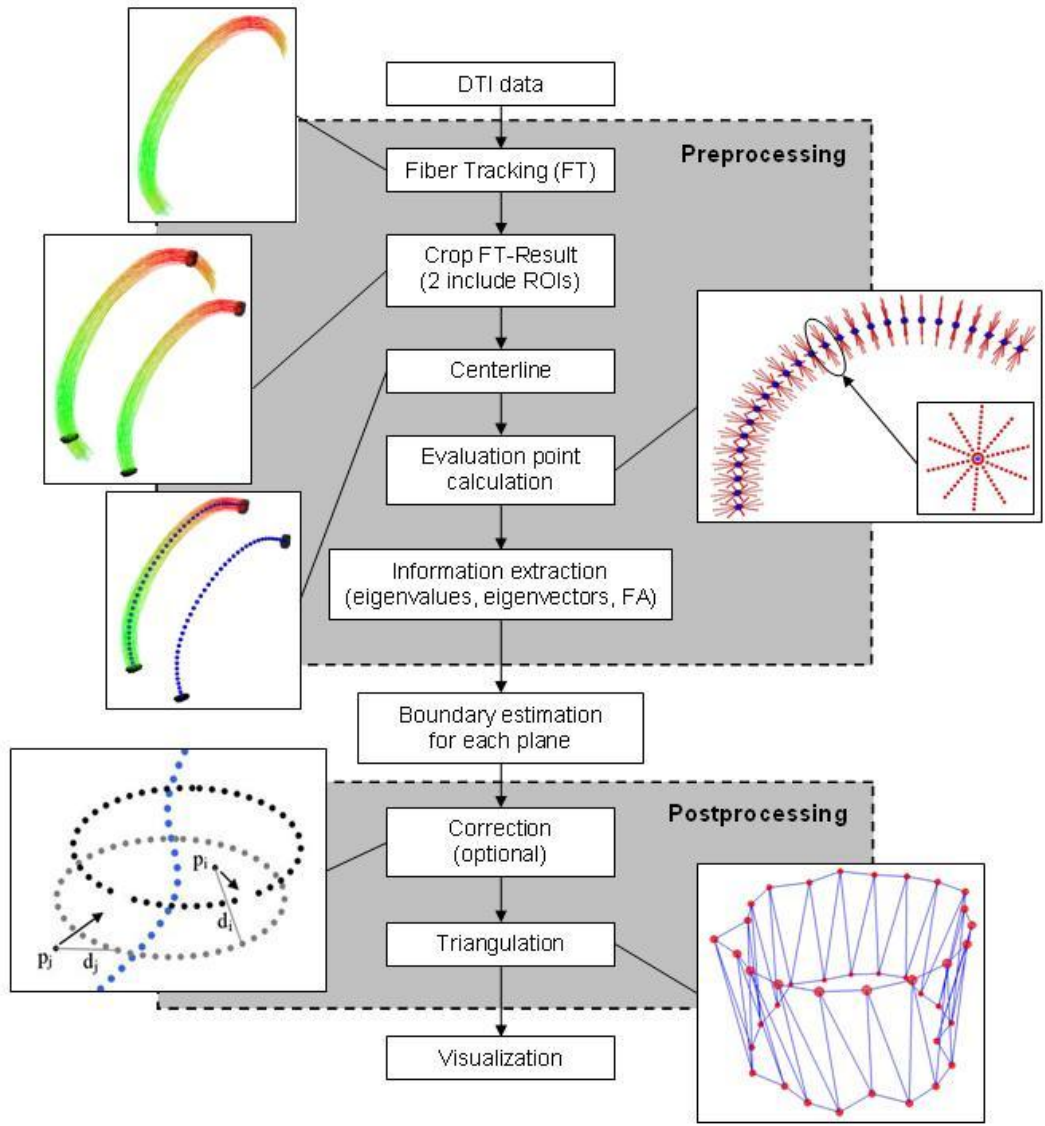

**Figure 1** – Concept of ray-based approach for fiber bundle estimation.

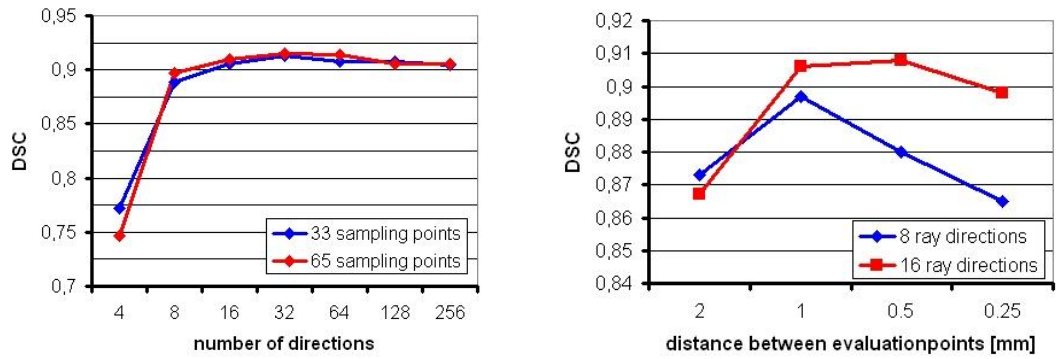

**Figure 2** – Dice Similarity Coefficient (DSC) with different number of sampling points and rays per layer (left) and DSC with different number of rays per layer and distances between evaluation points (right).